\documentclass{article}

\usepackage[nonatbib,preprint]{neurips_2019}

\usepackage[utf8]{inputenc} 
\usepackage[T1]{fontenc}    
\usepackage{hyperref}       
\usepackage{url}            
\usepackage{booktabs}       
\usepackage{nicefrac}       
\usepackage{microtype}      
\usepackage{ulem}

\usepackage{graphicx}
\usepackage{subfigure}
\usepackage{booktabs} 

\usepackage{bm}
\usepackage{amsmath,amssymb,amsfonts}
\usepackage{mathrsfs}

\usepackage[numbers]{natbib}

\usepackage{multirow}
\usepackage{lscape}
\usepackage{color}

\usepackage{wrapfig}

\title{Graph Warp Module: an Auxiliary Module for Boosting the Power of Graph Neural Networks in Molecular Graph Analysis}

\author{%
  Katsuhiko Ishiguro~~~~~Shin-ichi Maeda~~~~~Masanori Koyama \\
  Preferred Networks, Inc. \\
  Tokyo, Japan \\
  \texttt{k.ishiguro.jp@ieee.org, \{ishiguro,ichi,masomatics\}@preferred.jp} \\
}

\begin{document}

\maketitle

\begin{abstract}
Graph Neural Network (GNN) is a popular architecture for the analysis of chemical molecules, and it has numerous applications in material and medicinal science.
Current lines of GNNs developed for molecular analysis, however, do not fit well on the training set, and their performance does not scale well with the complexity of the network. 
In this paper, we propose an auxiliary module to be attached to a GNN that can boost the representation power of the model without hindering with the original GNN architecture. 
Our auxiliary module can be attached to a wide variety of GNNs, including those that are used commonly in biochemical applications.  
With our auxiliary architecture, the performances of many GNNs used in practice improve more consistently, 
achieving the state-of-the-art performance on popular molecular graph datasets.  
\end{abstract}

\section{Introduction}
\begin{figure}[t]
    \centering
    \includegraphics[width=124mm]{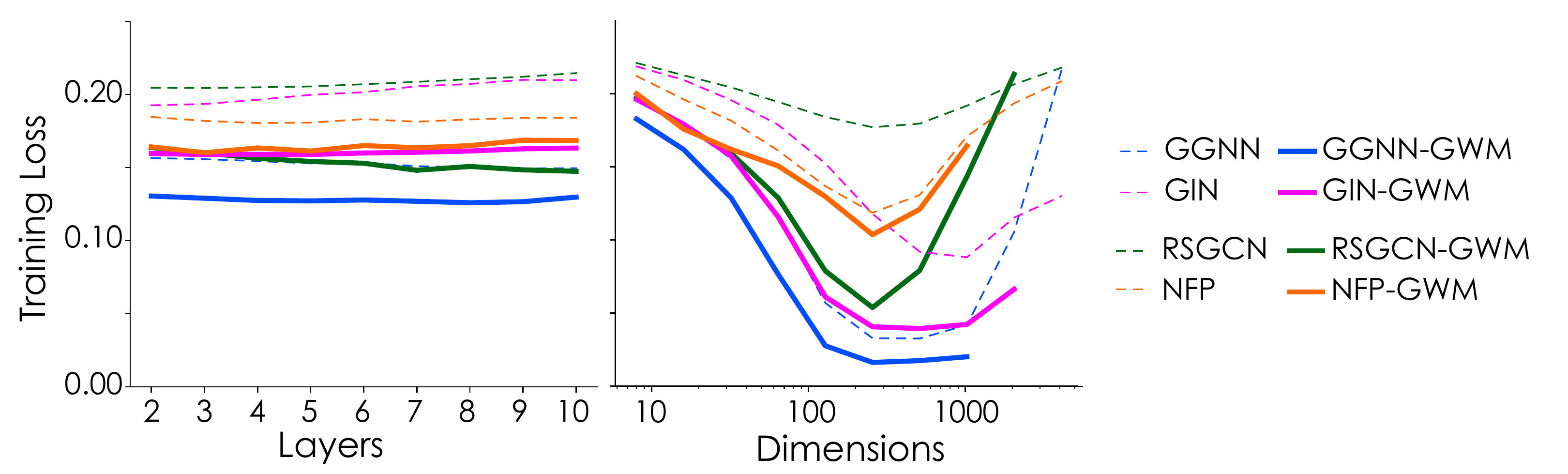}
    \caption{Training losses of various GNN models on a molecule graph dataset (Tox21). 
    The horizontal axis denotes the number of GNN layers (the left panel) or the dimension of the node feature vectors (the right panel). 
    Color denotes the GNN model. Thinner dashed lines are the losses of the vanilla GNNs, while thicker solid lines show the losses of the GNNs attached with the proposed Graph Warp Module (``GWM''). Scores are partially unavailable due to memory shortages. }
    \label{fig:result_layer_unit_stacking}
\end{figure}

Recently, Graph Neural Network (GNN) is a popular choice of model in the analysis of molecular datasets in medicinal and material science. 
Many molecular datasets consist of molecular graphs with feature vectors associated to each atom, and numerous methods based on GNN has been proposed to date just for learning the features of chemical molecules~\cite{Wu18MoleculeNet,Duvenaud15NIPS,Kearnes16,Li17arxiv,Gilmer17ICML,Shang18arxiv}, such as those pertaining to electrical conductivity and toxicity. 
One problem in the application of GNN to molecular datasets is the difficulty in reducing the training loss.
Unlike in the applications of Deep Neural Networks (DNNs) to image datasets, the training loss of GNN on molecular dataset does not decrease consistently with the number of layers nor number of nodes per layers (Fig.~\ref{fig:result_layer_unit_stacking}, thin dashed lines), and this seems to happen to numerous GNN architectures that are used in applications today~\cite{Duvenaud15NIPS,Li16ICLR,Kipf_Welling17ICLR,Xu19ICLR,Busbridge18OpenReview}. 
Unfortunately, many strong techniques developed for deep CNNs such as ResNet~\cite{He16CVPR} cannot be applied naively to GNN, because the tasks for GNNs are oftentimes fundamentally different in nature from that of standard DNN. 
For example, each graph data to be passed to the network can differ in size, and it is also often desired that GNN is equivariant (invariant under the reordering of vertices) in general. 
To the best of the authors' knowledge, there have not been many studies done to date that directly addressed the problem of training loss.

In this study, we propose graph warp module (GWM), a supernode~\cite{Li17arxiv,Gilmer17ICML,Battaglia18arXiv} based auxiliary module that can be attached to generic GNNs of various types to improve its representation power. 
Thick solid lines in Fig.~\ref{fig:result_layer_unit_stacking} plot the training losses of networks augmented with GWM. 
In general, GWM has an effect of decreasing the training loss for most choices of numbers of layers and dimensions of the node feature vectors, for all GNNs.
GWM is \textit{modular} in that its I/O is defined independently from the GNN to which it is attached, and the user can improve the representation power of the original GNN just by adding a small segment of code to add the module.

Our GWM is a combination of existing architectures, 
and it consists of three major components. 
The first component is \textit{virtual supernode} \cite{Li17arxiv,Gilmer17ICML}, which communicates with \textit{all} nodes in the graph and promotes the remote message passing. 

%
The second and third components are attention unit ~\cite{Vaswani17NIPS_AttentionIsAllYouNeed,Velickovic18ICLR} and gating units\cite{Cho14EMNLP}. 
These adaptive weighting functions in the module help adjust the flow of messages and deliver a message of appropriate strength to each node in the graph. 

Our contributions are as follows: 
\begin{enumerate}
    \item We introduce GWM, an auxiliary module that can help improve the representation power of the GNNs that are designed for the analysis of small graphs. 
    \item We show that, with a GWM, we can create a GNN that has state-of-the-art generalization performance on popular molecular graph datasets. 
\end{enumerate}

\section{Related Work}


\subsection{Virtual supernode}


A common challenge in the application of GNNs to a graphical dataset is the difficulty in propagating the information across remote parts of graphs.
Previously proposed solutions include sub-sampling~\cite{Hamilton17NIPS} and pooling of neighbor nodes~\cite{Ying18NeurIPS}. 
However, these clustering approaches are not too effective on the analysis of small graphs, in which every  node can have a strong influence on the graph label.

In this study, we use supernode~\cite{Gilmer17ICML,Li17arxiv,Pham17arxiv} to promote the global propagation of the information in molecular graphs.
By adding a supernode to a graph, we can allow any pair of nodes in the graph to communicate through the supernode in one hop. 
Battaglia et al \cite{Battaglia18arXiv} discusses a framework of GNNs that generalizes the supernode-augmented GNNs. 
One advantage of the supernode-based approach is that we can modify the network architecture while keeping the original GNN model intact. 
However, naive addition of a supernode to a graph can potentially lead to inadvertent over-smoothing of information propagation (c.f.~\cite{Li18AAAI}). 
In our study, we therefore make the supernode a \textit{module} by combining it with a \textit{gated message passing mechanism}.
This auxiliary module enables us to regulate the amount and type of information that is propagated through the feature space of the supernode.

\subsection{Message Passing and Attention/Gate Mechanism in GNN}
The supernode in our GWM transmits information using the mechanism of  message passing neural network (MPNN)~\cite{Gilmer17ICML}, which is defined 
recursively as follows  by composing multiple layers of the form: 
\begin{align}
h_{\ell, i} =  \mathscr{F}_\ell \left(\{h_{\ell-1, j}; j \in N(i)\cup\{i\}\} \right) 
\end{align}
where $i,j$ are indices of nodes in a graph. $h_{\ell, i}$ is the feature vector of the node $i$ at the $\ell$th layer, 
$N(i)$ is the neighborhood of the node $i$, 
and $\mathscr{F}_\ell$ is an appropriate choice of function that updates the feature vectors of the previous layer.
That is, MPNNs work by passing the information of each node to its neighbors in a recursive manner. 
Various methods are proposed for the choice of $\mathscr{F}$ and for the method of pooling the information of the neighbors of each node~\cite{Schlichtkrull17arxiv,Kipf_Welling17ICLR,Li16ICLR,Bruna14ICLR,Duvenaud15NIPS}.

Attention mechanism is a mechanism that helps the network regulate the importance of each node/edge in message passing (cf.~\cite{Wang18CVPR}). 
A Relational GCN~\cite{Schlichtkrull17arxiv} assumes that the aggregation weights of $h_{\ell -1,j}$ is fixed a priori in all $\mathcal{F_\ell}$.
With such architectures, however, one cannot regulate the higher order correlation amongst the outputs from each node.
Graph Attention Networks (GATs) \cite{Velickovic18ICLR} 
introduce a self-attention mechanism~\cite{Vaswani17NIPS_AttentionIsAllYouNeed}, which is equipped with a trainable set of weights that controls the importance of edges for each node. 
The relational graph attention network (RGAT)~\cite{Busbridge18OpenReview} also builds upon GAT
and constructed multiple types of attentions derived from relation-type-wise intermediate node representations. 
Finally, a GRU~\cite{Cho14EMNLP} is a gating mechanism originally introduced for recurrent neural networks. 
Gated Graph sequence Neural Networks (GGNN)~\cite{Li16ICLR} are the first to apply GRUs to the GNNs, and their method aims to introduce a recurrence relation between successive layers.
Our GWM is equipped with both multi-relational attention mechanisms and GRUs to grant the module greater flexibility in the transmission of messages to the graph nodes.

\section{Graph Warp Module}

\begin{figure}[t]
\centering
\includegraphics[width=139mm]{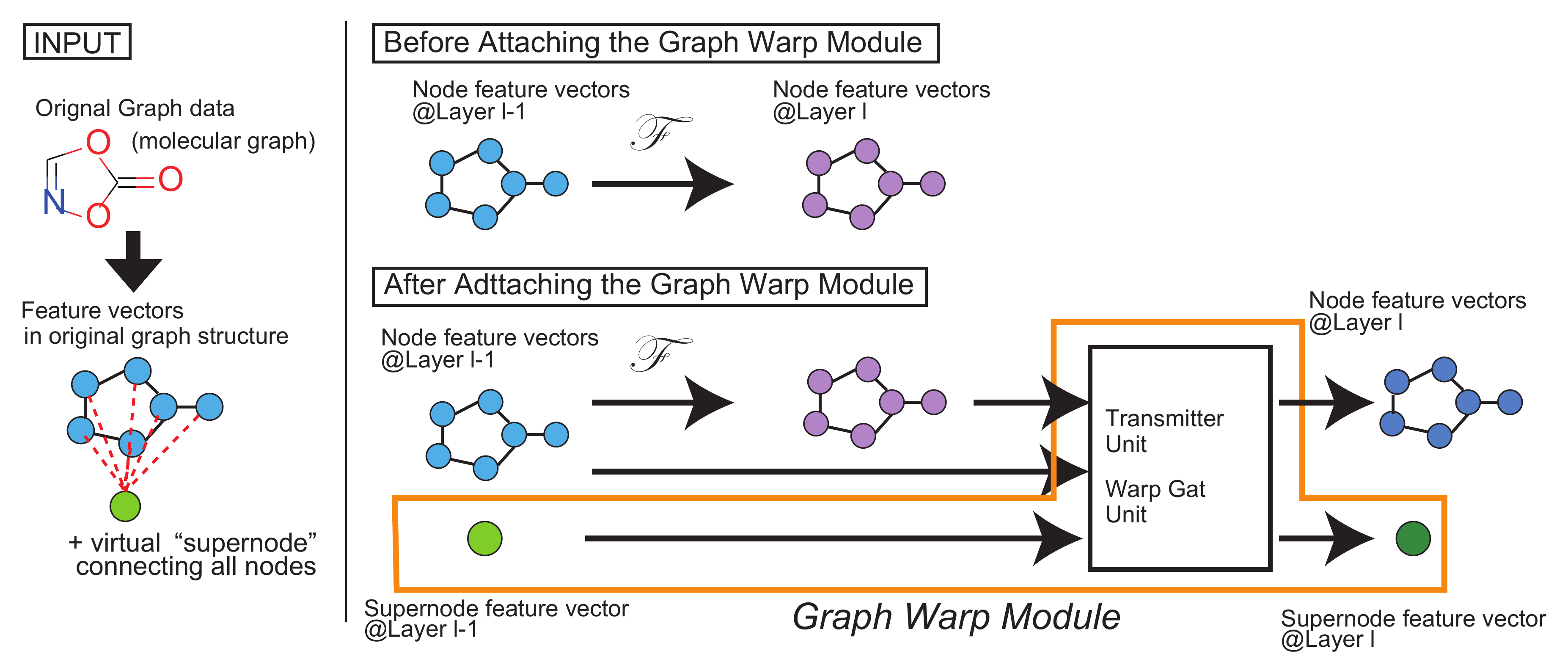}
\caption{
The overview of the proposed Graph Warp Module (GWM). 
A GWM consists of a supernode, a transmitter unit, and a warp gate unit. A GWM can be added to the original GNN as an auxiliary module.
At each layer, the supernode and the main network communicate through the transmitter and the warp gate. 
}
\label{fig:model_overview}
\end{figure}

Our Graph Warp Module (GWM) is made of three building blocks: 
(1) a supernode, (2) a transmitter unit, and (3) a warp gate unit (Fig.~\ref{fig:model_overview}).    
In a GWM-attached GNN, information is propagated across the graph through \textit{communication} between the supernode and the original (main) GNN at each layer. 
Messages from the supernode and the main GNN are transmitted to the warp gate through the transmitter unit, 
and the results of the communication are passed back to the module/main network through the warp gate units. 
In this section, we describe the Graph Warp Module in detail, and present the motivation of our design.

\subsection{Premise: vanilla GNN and its I/O}
Before describing our GWM, we need to present the I/O notation for the family of GNNs we consider, and explain how they will be used when a GWM is attached to a GNN.
We denote an arbitrary graph with the edge set $E$ and the node set $V$ as $G=(V,E)$.  
We will label the nodes in $V$ as $i=1,2,... |V|$, and represent each edge as a pair of nodes in $V$. 
The adjacency matrix $\mathcal{A} \in \mathbb{R}^{|V| \times |V|} $ is a matrix whose $(i,j)$th entry is the weight assigned to the edge between the node $i$ and the node $j$. 
Each instance of \textit{data} passed to the GNN is a set of input feature vectors. 
We denote an input feature vector associated with node $i$ as $x_{i}$. 
The type of GNN that we consider computes the output recursively by applying a composition of smooth functions $\mathscr{F}_{\ell}$ to $x_i$s.  
With the understanding that $x_j = h_{0,j}$,  
let $h_{\ell, i} = \mathscr{F}_{\ell-1, i}(h_{\ell-1,j} ; j \in V) \in \mathbb{R}^d$ be the vector of features to be assigned to the $i$th node by the $\ell$th layer of the GNN.
When the GNN is operating on its own without the attachment of a GWM, the GNN updates a feature vector using $h_{\ell, i} = \mathscr{F}_{\ell-1, i}(h_{\ell-1,j} ; j \in V) \in \mathbb{R}^d$. 
Finally, the GNN reports some form of the aggregation of 
$\{h_{L,i}; i \in V\}$ as the final Readout output. 

When a GWM is attached to the GNN, the main(bulk) GNN is requested to report $\mathscr{F}_{\ell-1, i}(h_{\ell-1,j} ; j \in V) \in \mathbb{R}^d$ as the message from the $\ell-1$th main layer to the module, where it is treated as an element in the \textit{intermodule hyperspace} and is mixed with the transmission from the supernode. 
The GWM will return the mixed message $h_\ell$ back to the $\ell$th layer of the main GNN.
At the same time, the GWM requests a transmission message from the main GNN to the $\ell$th supernode. 
The module will mix the transmission and the message from the $\ell-1$th supernode and return the mixed message $g_\ell$ to the $\ell$th supernode.  
The final output is produced by aggregating $\{h_{L,i}; i \in V\}$ and $g_L$.

\subsection{Supernode}
A supernode is a special node that is connected to nodes in the original graph to 
promote global information propagation across the network (Fig.~\ref{fig:model_overview}).
A supernode is to be prepared for each $\ell$th layer of the main GNN, and we associate a feature vector $g_\ell$ to the supernode at the $\ell$th layer. 
At each layer, the transmitter requests the following from the supernode: (1) a message $\mathscr{G}_{\ell} \left( g_{\ell} \right)$ for the $\ell+1$th layer and (2) a transmission to the main network, where $\mathscr{G}_\ell$ is an appropriate choice of a smooth function.  

Because a supernode is a superficial variable, we must initialize $g_0$ manually. For instance, we can use some form of aggregation of the global graph features (e.g. a number of nodes or edges, graph diameter, girth, cycle number, min, max, histogram, or an average of input node features, $\dots$). 
A detailed example of the aggregated feature is presented in the appendix. 

\subsection{Transmitter Unit}
\begin{figure}[t]
\centering
\includegraphics[width=139mm]{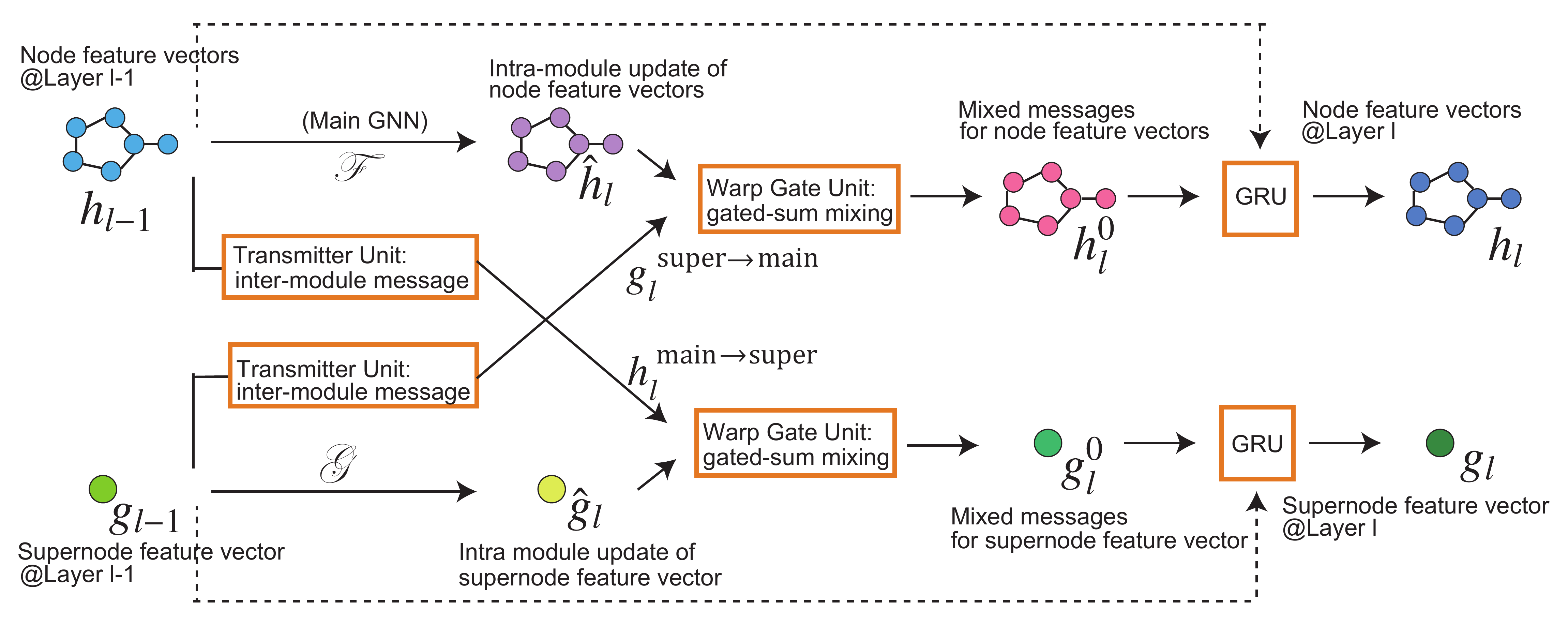}
\caption{
Details of the GWM computations. 
}
\label{fig:transmitter_mergergate}
\end{figure}

The transmitter unit handles the communications between the main GNN module and the GWM (Fig.~\ref{fig:transmitter_mergergate}). 
The transmitter module is responsible for translating the messages from the recipient
into a form that can be mixed in the intermodule hyperspace. 
We will use multiple types of messages and thus use a separate attention mechanism for each type of message. 
Before transmitting messages from the main GNN to the supernode, the transmitter uses the $K$-head attention mechanism to aggregate messages of each type. 
We enumerate the components included therein: 
\begin{itemize}
    \item $m_{\ell, k}^{\textrm{main} \to \textrm{super}} $:  aggregated message of head $k$ from the main GNN to the supernode at layer $\ell$. 
    \item $h_\ell^{\textrm{main} \to \textrm{super}}$: transmission from the main GNN to the supernode, derived from $m_{\ell, k}^{\textrm{main} \to \textrm{super}} $.   
    \item $g_{\ell}^{\textrm{super} \to \textrm{main}}$: transmission from the supernode to the main at layer $\ell$.
\end{itemize}
The transmissions are to be constructed from the following set of equations.
For a vector $v$, we use $v_{m:n} \in \mathbb{R}^{(m-n)d}$ to denote the concatenation of the vectors $v_m, v_{m+1},... \in \mathbb{R}^d$. 
Throughout, we use capital letters to denote the trainable coefficients.
\begin{equation}
	h_{\ell}^{{\textrm{main} \to \textrm{super}}} = \tanh \left(
	W_{\ell} ~  m_{\ell, 1:k}^{\textrm{main}\to \textrm{super}} 
    \right) \in \mathbb{R}^{D'} \, , 
    \label{eq:GWM_h_trans_l}
\end{equation}
\begin{equation}
	m_{\ell, k}^{\textrm{main}\to \textrm{super}} = 
    	\sum_{i} \alpha_{\ell, i, k} U_{\ell,k} h_{\ell-1, i}
    \in \mathbb{R}^{D'} \, , 
    \label{eq:GWM_m_lsk}
\end{equation}
\begin{equation}
	\alpha_{\ell, i, k} = \text{softmax} \left( h_{\ell-1,i}^{T} A_{\ell,k} g_{\ell-1} \right)  \in (0,1) \, , 
    \label{eq:GWM_alpha_lik}
\end{equation}
where $\alpha_{\ell, i, k}$ denotes an attention weight of the $i$th node at the $k$th head (type) and the $l$th layer.  

The transmission from the supernode to the main is simply given by:
\begin{equation}
	g_{\ell}^{\textrm{super} \to \textrm{main}} = \tanh \left( F_{\ell} g_{\ell-1} \right) \in \mathbb{R}^{D}  . 
    \label{eq:GWM_g_trans_l}
\end{equation}
There is no analogue of $m$ for the supernode because we are not considering a set of messages of different types to be transmitted from the supernode. 

\subsection{Warp Gate} 
The warp gate is responsible for merging the transmitted messages and for passing the results to the supernode and the main network through self recurrent units.  
The gate uses \textit{warp gate coefficients} to control the mixing-rate of the messages. 
The components of the warp Gate are:
\begin{itemize}
    \item $h_{\ell}^0$: inputs to the GRU unit at $\ell$th layer that transmits the message to the main GNN. 
    \item $g_{\ell}^0$: inputs to the GRU unit at $\ell$th layer that transmits the message to the supernode. 
    \item $\hat h_{\ell, i}$: the message $\mathscr{F}_{\ell-1, i}(h_{\ell-1,k} ; k \in V) \in \mathbb{R}^{D}$ from the $\ell-1$th layer of the main network, expressed in the intermodule hyperspace. 
    \item $\hat g_{\ell}$: 
    the message $\mathscr{G}_{\ell-1} \left( g_{\ell-1} \right) \in \mathbb{R}^{D'}$ from the $\ell-1$th supernode, where $\mathscr{G}$ is an appropriate smooth function with outputs in the intermodule hyperspace.
    \item $z_{\ell, i}$: tensor of warp gate coefficients 
    for the transmission from the supernode to the main GNN.
    \item $z_{\ell, i}^{(S)}$: tensor of warp gate coefficients
    for the transmission from the main GNN to the supernode. 
\end{itemize}
The module then mixes the transmissions and the messages from the previous layer by applying the following gated interpolations: 
\begin{equation}
    h_{\ell,i}^{0} =  
    (1 - z_{\ell,i}) \odot \hat{h}_{\ell-1, i} + z_{l,i} \odot g_{\ell}^{\text{super} \rightarrow \text{main}}
    \in \mathbb{R}^{D} \, , 
    \label{eq:GWM_h_merge_li}
\end{equation}
\begin{equation}
    g_{\ell}^{0} = 
    z_{\ell}^{(S)} \odot h_{\ell}^{\text{main} \rightarrow \text{super}} + (1 - z_{\ell}^{(S)}) \odot \hat{g}_{\ell}
    \in \mathbb{R}^{D'} \, , 
    \label{eq:GWM_g_merge_l}
\end{equation}
\begin{equation}
	z_{\ell,i} = \sigma \left(
    	H_{\ell} \tilde{h}_{\ell, i} + G_{\ell} g_{\ell}^{\text{super} \rightarrow \text{main}}
    \right)  \, , \quad 
    z_{\ell}^{(S)} = \sigma \left(
    	H^{(S)}_{\ell} h_{\ell}^{\text{main} \rightarrow \text{super}} 
    	+ G^{(S)}_{\ell} \hat{g}_{\ell}
    \right) \, , 
    \label{eq:GWM_z_li_ls}
\end{equation}
where $\sigma$ is a nonlinear function whose range lies in $[0, 1]$. 
Finally, the warp gate returns the mixed messages to the main GNN and the supernode through gated recurrent units:
\begin{equation}
    h_{\ell,i} =  
    \textbf{GRU} \left( h_{\ell-1,i}, h_{\ell,i}^{0} \right)
    \in \mathbb{R}^{D} \, , 
    \quad 
    g_{\ell} = \textbf{GRU} \left( g_{\ell-1}, g_{\ell}^{0} \right)
    \in \mathbb{R}^{D'} \, . 
    \label{eq:GWM_h_li_g_l}
\end{equation}

As we will show with ablation studies (Sec. 4.3), every component of GWM is essential in making the module work. 
The Attention coefficients  (Eqs.(\ref{eq:GWM_m_lsk},\ref{eq:GWM_alpha_lik})) are important because the amount and the type of information that must be transmitted to remote nodes may differ for different nodes. 
The Gating coefficients (Eqs.(\ref{eq:GWM_h_merge_li}-\ref{eq:GWM_z_li_ls})) are important because we want to regulate the transmission from each node in the graph to the supernode and vice versa.  
We use different recurring units (Eq.\ref{eq:GWM_h_li_g_l}) for the transmission from the module to the supernode and the transmission from the module to the main network for each layer because the amount of the information that must be reinforced may differ for the main network and the supernode at each layer.

\section{Experiments}

In this section, we present our experimental results on multiple molecular graph datasets, 
testing the efficacy of the GWM for graph regression tasks and graph classification tasks.   

\subsection{Datasets}

We used four datasets collected in MoleculeNet~\cite{Wu18MoleculeNet}. 
These datasets are described in the SMILES string format, which admits the graph representations we described above.
For details, please see~\cite{Wu18MoleculeNet}.

For the graph regression tasks, we used the QM9 dataset and the Lipophilicity (LIPO) dataset.
QM9 is a dataset with numerical labels, containing about 133K drug-like molecules with 12 important chemical-energetic, electronic, and thermodynamic properties, such as HOMO, LUMO, and electron gaps.
The LIPO dataset is another numeric-valued dataset, containing the solubility values of roughly 4K drug molecules.
Each instance of data in these datasets is a pair of a molecular graph and a numerical value(s): the 12 chemical properties in the QM9 dataset, and the solubility in the LIPO dataset.
For both datasets, the task is to predict the numerical value(s) from the molecular graph. 
We evaluated the performance of the models using mean absolute errors (MAEs).  

For the graph classification tasks, we used the Tox21 and the HIV datasets. 
The Tox21 dataset contains about 8K pairs of molecular graph and 12 dimensional binary vector that represent the experimental outcomes of 
toxicity measurements on 12 different targets. 
The HIV dataset contains roughly 42K pairs of molecular graph and binary label that represent the medicinal effect of the molecule. 
For these datasets, the task is to predict the binary label(s) from the molecular graph. 
For these tasks, we use ROC-AUC values as a measure of performance. 

Throughout, we used the train/validation/test data splits of the ``scaffold'' type, which is considered by~\cite{Ruddigkeit12QM9,Ramakrishnan14QM9} as the difficult type for test predictions. 
Please find the appendix for details. 

\subsection{Choices of the Main GNN Models and Implementations}

We test GWMs on various GNN models. 
Neural Fingerprints (NFP)~\cite{Duvenaud15NIPS} and Weavenet~\cite{Kearnes16} 
are relatively classical baselines. 
A Gated Graph Neural Network (GGNN)~\cite{Li16ICLR} is a strong GRU-based GNN. 
Renormalized Spectral Graph Convolutional Network (RSGCN\footnote{Referred as RSGCN in Chainer-Chemistry package, but often simply referred as ``GCN'' in several papers.})~\cite{Kipf_Welling17ICLR}, a popular GNN model approximating a CNN for graphs~\cite{Defferrard16NIPS}. 
The relational graph attention network (RGAT)~\cite{Busbridge18OpenReview} uses multiple attention 
mechanisms for a set of edge types. 
Graph Isomorphism Network (GIN)~\cite{Xu19ICLR} employs multi layer perceptrons within each layer for richer transformations. 

We implement all models in Chainer~\cite{Chainer}.
In the readout layer, we first aggregate all information from the main nodes in the same way as in the original paper, concatenated the result with the features from the supernode, and passed the concatenated tensor to a fully connected layer. 
For evaluation, we used a softmax cross entropy for classification tasks and a mean squared error for regression tasks.
We use a fixed set of hyperparameters throughout the study.

%
All models were trained with Adam~\cite{Kingma_Ba15ICLR}. 
We report the results of the model snapshots of the epoch for which the best validation score was achieved. 
For implementation details including readouts and hyperparameters, please read the appendix.

\subsection{Training and Test Loss Reduction}

A brief peek of the results in the introduction shows that the installment of GWM can improve the representation power of the GNN for the classification task on the Tox21 dataset.  

In the first experiment, we studied the effect of the GWM on the training loss and the test loss for all 16 pairs in \{4 datasets\} $\times$ \{4 GNNs (used in Fig.~\ref{fig:result_layer_unit_stacking})\}.
We report the average $\bar r$ of the loss reduction ratio 
$r = \frac{\mathscr{L} - \mathscr{L}^{(+)}}{\| L \|}$
for both training loss and test loss over 10 runs. 
$\mathscr{L}$ denotes the loss of the vanilla model, and $\mathscr{L}^{(+)}$ denote the loss of the GWM-installed model.
$\bar r_{train}$ denotes a reduction ratio of the training loss, and $\bar r_{test}$ denotes a reduction ratio of the test loss. 

\begin{wrapfigure}{r}{77mm}
\centering
    \includegraphics[width=73mm]{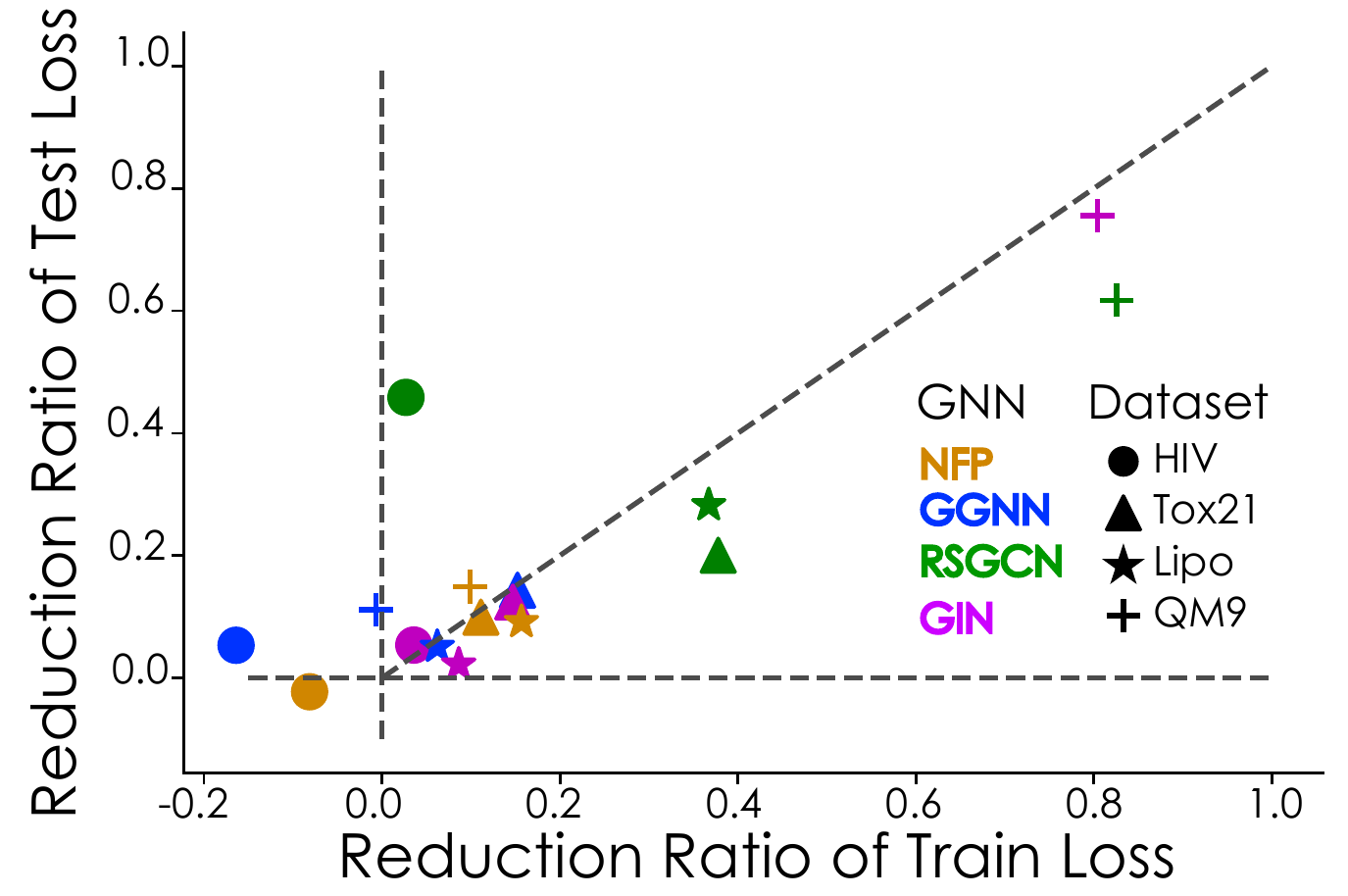}
    \caption{Train (horizontal) and test (vertical) loss reduction ratios on various pairs of GNN models and datasets. 
    Each plot presents the rational train/test loss reductions induced by the GWM attachment for a specific pair of (dataset(symbol), GNN(color)). 
    }
    \label{fig:result_loss_reduction_L3D50}
\end{wrapfigure}

Fig.~\ref{fig:result_loss_reduction_L3D50} is the scatter plot of the $(\bar r_{train},  \bar r_{test})$, with the dotted slope representing $\bar r_{train} = \bar r_{test}$. For all GNNs and datasets, we set $L=3$ and $D=50$.    
As we can see in the plot, $\bar r_{train}$s were negative for the two GNNs in the HIV dataset (blue and brown circles). $\bar r_{train}$ for GGNN on QM9 (blue cross) was a very small negative value). 
For the other 13 (model-dataset) pairs, the attachment of GWM consistently reduces the training loss( i.e. improves the fit to the training graph datasets. )
Remarkably,  15 out of 16 pairs had positive $\bar r_{test}$s: the GMW improved generalization performances in most cases.  
It is worthy of note that $\bar r_{train}$ and $\bar r_{test}$ are positively correlated in this scatter plot. 
We were able to obtain similar results for all other choices of hyperparameters we tested.
For the results with different hyperparameter values, see the appendix.
%
This result implies that our GWM has the general effect of improving the generalization performance by augmenting the representation power of the main GNN.

\subsection{Effect of the GWM on the representation power of model space}

\begin{table}[t]
    \centering
    \begin{tabular}{l||c|c|c}
         Model Name (emulating) & Attention & Gatings & GRUs \\ \hline
         Simple supernode (\cite{Li17arxiv}) & --- & --- & ---  \\
         NoGate GWM (\cite{Gilmer17ICML}) & $\checkmark$ & --- & $\checkmark$ (supernode only)  \\
         Proposed GWM & $\checkmark$ & $\checkmark$ & $\checkmark$  
    \end{tabular}
    \caption{Supernode-based models validated in the Experiment 4.4}
    \label{tab:compared_models}
\end{table}

In the second experiment, we studied the effect of GWM on the representation power of GNN models.
To compare the GWM-augmented GNNs with their vanilla GNN counterparts on fair grounds, 
we used Bayesian optimization to optimize the number of layers and the dimension of feature vectors for each model-dataset pair we tested.  
Bayesian optimization was conducted using a package in Optuna~\cite{Optuna}.

We conducted a set of ablation studies to investigate the effect of (1) attention mechanism, (2) Gating mechanism, and (3) the Recurrent unit.
We used two ablation models. 
\textbf{Simple supernode} model is a supernode without attention, gatings, and GRU functions.
This model is akin to the supernode model in~\cite{Li17arxiv}, except that our model allows bi-direction communication between supernode and the main network. 
\textbf{NoGate GWM} is a GWM without gatings, and it lacks GRU for the main GNN. 

Table~\ref{tab:compared_models} summarizes the resits of our ablation studies, 
For the detailed formulations of the two ablation models,  please see the appendix.

Table \ref{tab:result_lipo_qm9_BO} and Table \ref{tab:result_hiv_tox21_BO} respectively present the MAEs on the regression tasks and the ROC-AUCs for the classification tasks, averaged over 10 random runs. 
In the tables, bold faces indicate the improvements from the vanilla GNN and asterisks indicate the best model among supernode models for each (dataset, GNN) pair.  
Full tables with standard deviations are presented in the appendix. 
As we can see in the tables, the proposed GWM improves the generalization performances for 23 out of 24  (model-dataset) pairs.
These results suggest that the proposed (full) GWM can improve GNNs' performances irrespective of the choice of GNN models and the dataset. 

A few words of caution are in order here.
Two ablation models did not improve the generalization of GNNs for the QM9 and the HIV datasets (see the column ``\# Improved''). 
This suggests an appropriate combination of attentions, gatings, and GRUs is essential in making the supernode effective for the analysis of molecular graph datasets.


Most studies on the MoleculeNet present the performance of their methods on the random train/validation/test data split, 
despite the biochemical significance of the scaffold split. 
As an exception, Li et al.~\cite{Li17arxiv} report the test AUC scores on the Tox21 and the HIV\footnote{In~\cite{Li17arxiv}, the authors proposed to use the focal loss instead of the standard cross entropy for the HIV, which yields a better AUC than the vanilla GNN.} in the scaffold data split. 
The AUC score of~\cite{Li17arxiv} was $0.759$ for Tox21. 
The RGAT+GWM model outperformed this result ($0.787$.)
Meanwhile, \cite{Li17arxiv} and GGNN+GWM achieved similar result for HIV($0.763$ and $0.762$, respectively).
We also examined the performance-gain from the addition of GWM in terms of AUC. 
The performance gain achieved by~\cite{Li17arxiv} was $0.007$ for Tox21, and  $-0.006$ for HIV. 
Meanwhile, the performance gain achieved by GWM was $0.023$ for Tox21 (RGAT+GWM) and $0.018$ for HIV (GGNN+GWM). 

\begin{table}[t]
    \centering
    \small
    \begin{tabular}{c|c||c|c|c|c|c|c||c}
         Dataset & GNN model & NFP & Weave & RGAT & GGNN & RSGCN & GIN 
         & \# Improved\\ \hline 
         \multirow{2}{*}{LIPO} & vanilla GNN & .677  & 1.19 & .753 & .582  & .801 & .844 & - \\
                               & $+$Simple Supernode & 
                               .693  & 
                               \textbf{1.01}  & 
                               \textbf{.740}  & 
                               .604 & 
                               \textbf{.775} & 
                               \textbf{.819} & 4/6 \\
                               & $+$NoGate GWM & 
                               \textbf{.675}  & 
                               \textbf{.721}  & 
                               \textbf{.688}  & 
                               \textbf{.576}  & 
                               \textbf{.787} &  
                               .847 & 5/6 \\
                               & $+$Proposed GWM & 
                               \textbf{.672*} & \textbf{.688*} & \textbf{.659*} & \textbf{.569*} & \textbf{.752*}  & \textbf{.784*} 
                               & \textbf{6/6} \\
                               \hline
         \multirow{2}{*}{QM9} & vanilla GNN & 6.16  & 6.38 & 8.96 & 4.92  & 15.2  & 14.0 & - \\
                                & $+$Simple Supernode & 
                                7.68  & 
                                \textbf{5.51} & 
                                9.00 & 
                                5.41 & 
                                \textbf{14.6} &
                                \textbf{11.5*} & 3/6 \\
                               & $+$NoGate GWM & 
                               6.84  & 
                               \textbf{5.40*}  & 
                               9.21 & 
                               5.52 & 
                               \textbf{12.5} & 
                                \textbf{12.9} & 3/6 \\
                              & $+$Proposed GWM & 
                              6.64*& \textbf{5.90} & 
                              \textbf{8.39*}  & 
                              \textbf{4.88*} & \textbf{11.9*} 
                              & \textbf{11.8} & \textbf{5/6} \end{tabular}
    \caption{MAEs on the LIPO dataset and QM9 dataset. 
    Smaller values are better. 
    }
    \label{tab:result_lipo_qm9_BO}
\end{table}
\begin{table}[t]
    \centering
    \small
    \begin{tabular}{c|c||c|c|c|c|c|c||c}
         Dataset & GNN model & NFP & Weave & RGAT & GGNN & RSGCN & GIN 
         & \# Improved \\ \hline 
         \multirow{2}{*}{HIV} & vanilla GNN & .724 & .670 & .707 & .746 & .746  & .729 & - \\
                                & $+$Simple supernode & 
                                .707 & 
                                \textbf{.676} & 
                                .704  & 
                                \textbf{.764*}  & 
                                .728 & 
                                .729 & 2/6 \\
                               & $+$NoGate GWM & 
                               .714 & 
                               \textbf{.680} & 
                               \textbf{.726} & 
                               .744 & 
                               .742 & 
                               \textbf{.739} & 3/6 \\
                               & $+$Proposed GWM & 
                               \textbf{.731*} & 
                               \textbf{.681*} & \textbf{.748*} & \textbf{.762}  & \textbf{.758*} & 
                               \textbf{.755*} & \textbf{6/6}\\ \hline
         \multirow{2}{*}{Tox21} & vanilla GNN & .763 & .710 & .764  & .757  & .760 & .740 & - \\
                        & $+$Simple supernode & \textbf{.770} & \textbf{.750} & \textbf{.787*} & \textbf{.790} & \textbf{.770*}  & \textbf{.763} & 6/6 \\
                        & $+$NoGate GWM & 
                        \textbf{.775*} & 
                        \textbf{.764} & 
                        \textbf{.786}  & 
                        \textbf{.792*} & 
                        .759 &  
                        \textbf{.766} & 5/6 \\
                        & $+$Proposed GWM & 
                        \textbf{.769}  & 
                        \textbf{.767*} & 
                        \textbf{.787*} & 
                        \textbf{.785} & 
                        \textbf{.766} & 
                         \textbf{.768*} & \textbf{6/6} 
    \end{tabular}
    \caption{ROC-AUCs on the HIV dataset and Tox21 dataset. 
    Larger values are better. 
    }
    \label{tab:result_hiv_tox21_BO}
\end{table}

\section{Conclusion}

For a generic DNN, numerous effective \textit{installable modules} have been proposed for the improvement of the model
(e.g.~\cite{Srivastava14JMLR,Ioffe_Szegedy15ICML,Miyato18ICLR,He16CVPR}). 
The proposed GWM is the first of its kind to be installed to a generic GNN as an auxiliary module. 
Experimental results show that the GWM can generally improve the representation power as well as the generalization performance of a GNN, irrespective of the choice of GNN architecture and the molecular graph datasets. 
We would like to emphasize that the choice of the internal structure of GWM is not limited to the ones we described in this study, and that there are possibly numerous ways to construct a GWM-like module. 
For example, there is no provable justification for the use of a linear transformation in the transmissions or a bilinear form in the attention coefficients $\alpha$.  
Effective choices of supernode features are also open to further research. 
Our study can possibly open an entirely new avenue for the architectural study of GNNs.


\subsection*{Acknowledgement}
We authors would like to thank Dr. Prabhat Nagarajan for checking and revising the paper. 
We also thank Mr. Weihua Hu for his insightful comments about the experiments. 

\bibliographystyle{plain}
\bibliography{gwm}

\appendix

\section{Our formulation of RGAT}

Apart from the original RGAT~\cite{Busbridge18OpenReview}, 
we have developed a similar GNN in a slightly different formulation. 
Followings are our RGAT formulation: 
\begin{alignat}{2}
 h_{\ell+1,i} &= 
 \tanh \left( W_{l} \textbf{concat}_{k=1}^{K} \tilde{h}_{\ell,i,k} \right) \, , 
 \label{eq;h_l1i} 
 \\
 \tilde{h}_{\ell, i, k} &= 
 F_{\ell,k} h_{\ell, i} + \sum_{j \in N_{i}} \alpha_{i,j,k} G_{\ell,k} h_{\ell, j} \, , 
 \label{eq:tilde_h_lik} 
 \\
 \alpha_{i,j,k} &=
 \textbf{softmax} \left( 
 a \left( h_{\ell,i}, h_{\ell,j}; A_{\ell,k, e_{i,j}} \right) 
 \right) \, .
 \label{eq:alpha_ijk} 
 \\
  a \left( h_{\ell,i}, h_{\ell,j}; A_{\ell,k,e_{i,j}} \right) 
  &= 
 h_{\ell,i}^{T} A_{\ell,k,e_{i,j}} h_{\ell,j} \, . 
 \label{eq:a_ij}
\end{alignat}

$W, F, G, A$ are the coefficient matrix to be tuned. 
$\ell$ is the index of the layer up to $L$, 
$k$ is the index of the attention head up to $K$, 
$i,j$ are the index of the nodes up to $N$, 
$e_{i,j} = r$ is the index of the edge type up to $R$. 

The main point is the edge type information in Eq.\ref{eq:a_ij}. 
The edge type $e_{i,j} = r$ switches the weight matrix of the attention similarity function, $a$. 
This means that the associations between nodes should be computed dependent on the edge type. 
This is a natural assumption for chemical molecular graphs. Typically we have multiple bond types between nodes = atoms: single-bond, double-bound, triple-bond, and the aromatic ring. It is natural to assume that interactions between atoms are affected by the bond types among the atoms. 

The main differences from the original RGAT lie in the Eq.\ref{eq:tilde_h_lik}. 
The original RGAT assumes that the weight matrix $G$ is also dependent on the edge type ($G_{\ell,k,e_{i,j}}$) while we omit this dependency. 
Also, the original RGAT does not provide a self-link weight matrix $F$ while we do. 
We made these changes based on our preliminary experiments. We found our formulation is better than the original RGAT formulation in the MoleculeNet dataset, in terms of the training stability and the generalization performances. 

Another difference is the choice of the attention function. 
In our formulation, the attention similarity measure $a(\cdot)$ is defined by the \textit{general} attention in~\cite{Luong15EMNLP} while 
the original GAT~\cite{Velickovic18ICLR} and the RGAT~\cite{Busbridge18OpenReview} employed a simpler \textit{concat} attention. 

\section{Our Implementation of GIN}
We implement the simplest GIN: 2-layer MLP with ReLU activation for each layer and a bias parameter $\epsilon$ fixed at $0$. 
We regularize GIN with dropout~\cite{Srivastava14JMLR}, instead of batch-normalization~\cite{Ioffe_Szegedy15ICML}. 
This is because the batch-normalization of the Chainer-Chemistry library did not correctly treat the padded node elements in the minibatches when we conducted the experiments.

\section{Experiments Details: General Issues}

\subsection{Graph Data Representation}

All datasets used in our experiments are taken from the MoleculeNet\cite{Wu18MoleculeNet}. 
Four used datasets are provided in the SMILES format. 
A SMILES format is a line notation for describing the structure of chemical compounds. 
We decode a SMILES molecular data into a graph representation of the molecule. 
A node in the graph corresponds to an atom. Each atom node is associated with the symbolic label of the atom name (``H'', ``C'', ...). 
An edge in the graph corresponds to a bond between atoms. Each bond edge is associated with the bond type information (single-, double-, ....). 

Given the graph, we extract input feature vectors for node $x_{i}$ and that of supernode $x'$. 
$x_{i}$, the input feature vector for the node $i$ is a $D$-dimensional continuous vector, which is an embedded vector of the one-hot atom label vector with a trainable linear transformation. 
$X'$, The input feature vector for the supernode is a $D'$-dimensional continuous vector, which again is an embedded vector of some graph-global features with a trainable linear transformation. 
Choices for the graph-global features are detailed in the following section. 

The edge information is converted in an adjacency matrix, $\mathscr{A}$. 

\subsection{Explicit Features for Supernode}
Since the supernode does not exist in the original graph $G$, we have no observable cues for the supernode. 
For simplicity, we propose to use an aggregation of node features, such as: 
\begin{itemize}
    \item Histograms of discrete labels attached to original nodes
    \item Averages, maximums, minimums, or medians of numerical attributes attached to original nodes
    \item Histograms of edge types if the graph is multiple relational graph. 
    \item Number of nodes, graph diameters, modularity, and other simple statistics for graph structure. 
\end{itemize}
We can augment the super feature vector $x'$ if some additional information about the graph is provided. 
Essentially, these simple aggregations of the feature vectors do not bring new information into the network. However we found that the graph-wise super feature input 
boosts the performance of the learned network model. 

\subsection{Data splits}
In chemical datasets, a totally random shuffling of samples into train/val/test subsets is not always a valid way of data splitting. 
Therefore MoleculeNet provides several ways of data splitting. 
The ``random'' split is the random sample shuffling that are most familiar to the machine learning community. 
The ``scaffold'' split separate samples based on the molecular two-dimensional structure. Since the scaffold split separates structurally different molecules into different subsets, ``it offers a greater challenge for learning algorithms than the random split''~\cite{Wu18MoleculeNet}. 
Throughout the paper, we adopt the scaffold split to assess the full potential of the GWM-attaching GNNs. 

The actual construction of the scaffold split train/validation/test subsets has a freedom of algorithm choices. 
We basically adopted the algorithm provided by the deepchem\footnote{https://deepchem.io/} library, which is the standard split algorithm for many papers. 
However, for the experiment of train/test loss comparison, we adopted the algorithm provided by the Chainer Chemistry library. 

\subsection{Readout Layer}

In many applications of GNNs users may expect a single fixed-length vector representing the characteristics of the graph $G$. 
So we add the 'readout' layer to aggregate the original node hidden states $\{H_{\ell}\}$ and the global node hidden states $\{g_{\ell}\}$. 

A main issue in the readout unit is how to aggregate the original nodes, whose number varies for each graph. 
A simple way is to take an arithmetic average (sum) of the $h$s at the $L$-th layer, but we can also use a DNN to compute (non-linear) ``average'' of $h$s~\cite{Li16ICLR,Gilmer17ICML}.  
After the aggregation of the node hidden states, we simply concatenate it with $g$s and apply some transformations to 
achieve the readout vector, $r$: 
\begin{equation}
    r = \text{DNN}_{r1} \left(
    \text{concat} \left[ \text{DNN}_{r2} \left(H_{L}\right) , g_{L} \right]
    \right) \, . 
    \label{eq:GWM_readout}
\end{equation}
In the above equation, $\text{DNN}_{r1}$ is a multi-layer perceptron or a fully connected layer to mix the concatenated hidden vectors. We adopted a simple fully-connected layer for $\text{DNN}_{r1}$ in this paper. 
$\text{DNN}_{r2}$ is a specific readout unit accompanied with a original GNN to aggregate variable-length $H_{L}$. 

\subsection{Optimizer}
All models were trained with Adam~\cite{Kingma_Ba15ICLR}, $\alpha = 0.001$, $\beta_{1} = 0.9$, and $\beta_{2}=0.999$. 

\subsection{Ablation models formulation}
Here we detail the formulation of the ablation models used in the main comparison experiments (Sec. 4.4).

As written in the main manuscript, we formulate the two ablation models (Table \ref{tab:compared_models}) as follows. 
A \textbf{simple supernode} model, which slightly augments~\cite{Li17arxiv}, simplify all attentions, gates, and the GRUs. 
First, there is no attention for the Transmitter. So the message from the main nodes to the supernode is just a sum of hidden vectors, $h_{\ell-1,:}$: 
\begin{equation}
	h_{\ell}^{{\textrm{main} \to \textrm{super}}} = 
	\tanh \left(
	W_{\ell} ~  \sum_{i} h_{\ell-1, i} \right)
    \in \mathbb{R}^{D'} \, . 
    \label{eq:Li_h_l}
\end{equation}
Originally there is no messages from the supernode to the main GNN in~\cite{Li17arxiv}, but we allow such a simple message in this ablation model: 
\begin{equation}
	g_{\ell}^{\textrm{super} \to \textrm{main}} = \tanh \left( F_{\ell} g_{\ell-1} \right) \in \mathbb{R}^{D'}  . 
    \label{eq:Li_g_trans_l}
\end{equation}
Messages are merged by simple linear combinations, instead of gates and GRUs, following~\cite{Li17arxiv}: 
\begin{equation}
    h_{\ell,i} =  
    \bm{Z}_{\ell, 1} \hat{h}_{\ell, i} 
    + 
    \bm{Z}_{\ell, 2} g_{\ell}^{\text{super} \rightarrow \text{main}}
    \in \mathbb{R}^{D} \, , 
    \label{eq:Li_h_li}
\end{equation}
\begin{equation}
    g_{\ell} = 
    \bm{Z}_{\ell, 1}^{(S)} h_{\ell}^{\text{main} \rightarrow \text{super}} 
    + 
    \bm{Z}_{\ell, 2}^{(S)} \hat{g}_{\ell}
    \in \mathbb{R}^{D'} \, . 
    \label{eq:Li_g_l}
\end{equation}

We find it difficult to fully recover the supernode of~\cite{Gilmer17ICML} since their description on the supernode is quite limited. 
Thus, a \textbf{NoGate GWM} model, which surrogates~\cite{Gilmer17ICML}, only capture the essence of their supernode: no gatings for merger, and GRU is not installed for the nodes of the main GNN. 
In this model, we use the same attention-based Transmitter unit as in Eqs.(2-6). We reduce the adaptive gatings in the Warp unit by a simple averaging, and omit the GRU for $h_{\ell, i}$. 
\begin{equation}
    h_{\ell,i}^{0} =  
    \bm{Z}_{\ell, 1} \hat{h}_{\ell-1, i} 
    + 
    \bm{Z}_{\ell, 2} g_{\ell}^{\text{super} \rightarrow \text{main}}
    \in \mathbb{R}^{D} \, , 
    \label{eq:Gilmer_h_merge_li}
\end{equation}
\begin{equation}
    g_{\ell}^{0} = 
    \bm{Z}_{\ell, 1}^{(S)} h_{\ell}^{\text{main} \rightarrow \text{super}} 
    + 
    \bm{Z}_{\ell, 2}^{(S)} \hat{g}_{\ell}
    \in \mathbb{R}^{D'} \, , 
    \label{eq:Gilmer_g_merge_l}
\end{equation}
\begin{equation}
    h_{\ell,i} =  
    h_{\ell,i}^{0}
    \in \mathbb{R}^{D} \, , 
    \label{eq:Gilmer_h_li}
\end{equation}
\begin{equation}
    g_{\ell} = \textbf{GRU} \left( g_{\ell-1}, g_{\ell}^{0} \right)
    \in \mathbb{R}^{D'} \, .  
    \label{eq:Gilmer_g_l}
\end{equation}

\subsection{Hyperparameter}

We fix a part of hyperparameters throughout the experiments, which does not influence performances so much:
the number of heads in all multi-head attention mechanisms to $K=8$, and used $R=4$ edge types for the multi-relational mechanism in all models. 
Also, at every layer, we set the dimension of the supernode feature to be the same as that of the features of the nodes in the main GNN.

In the next section, we list the other hyperparameters (the number of layers $L$, the dimension of feature vectors $D (=D')$) used in several experiments/figures, as well other experimental/implementation details. 

\subsection{Computational Environment}
We use a single GPU (mainlly nvidia Tesla V100) for an experimental run. 
A run roughly takes 1 hour to 1 day, depending on the hyperparameters and the GNN models. 

\section{Experiments Details: for each experiment}

In this section, we report details for each experiment, including the chosen hyperparameters and additional results. 

\subsection{Figure 1: Training Loss over $L$ and $D$}
For the experiments on Figure 1 in the main manuscript, we used the Tox21 dataset. 

To study the effect of the number of layers $L$ (the left panel), 
we fixed the dimension $D=32$. 
To study the effect of the feature vector dimensions $D$ (the right panel), 
we fixed the number of layers $L=4$. 

All models are trained for 30 epochs. 

\subsection{Section 4.3: Train/Test Loss Reduction}
In the experiment in the Section 4.3 (Figure 4 in the main manuscript), we align the hyperparameters including $L, D$ among a vanilla GNN and its GWM-installed counterpart, to compare the loss function values. 
We used four datasets. For each dataset, we fixed $L$ and $D$ for all GNN models to compare the loss reduction performances. 

In Figure 4, $L=3$ and $D=50$ for all GNNs and datasets. 
In preliminary experiments, we manually changed $L$s ($\in \{2, 3, 4\}$) and $D$s ($\in 32, 50, 100, 150$) in some extent, but found the overall tendency of the scatter plots does not dramatically change. 
This is partially understood from the Figure 1 in the main manuscript: the loss curves of the vanilla GNNs and their GWM-augmented counterparts evolve in roughly parallel. This implies the ratios of loss reductions are not so much dependent on the hyperparameter choices. 

Here we show the results of other hyperparameter settings. All cases we observe the similar plot patterns. 

\begin{figure}[h]
\centering
    \includegraphics[width=73mm]{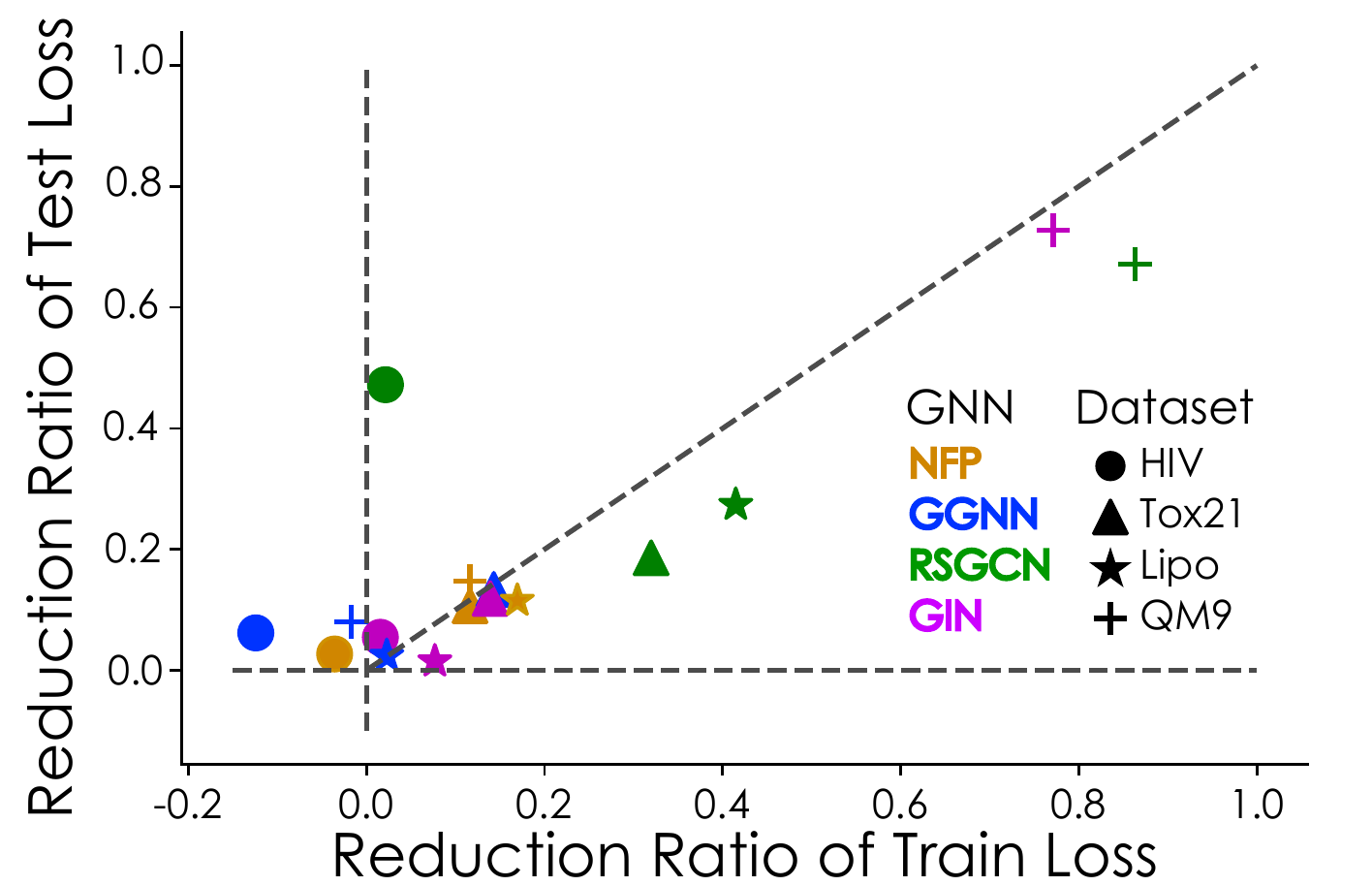}
    \caption{Train (horizontal) and test (vertical) loss reduction ratios on various pairs of GNN models and datasets. $L=3$, $D=32$. 
    Each plot presents the rational train/test loss reductions induced by the GWM attachment for a specific pair of (dataset(symbol), GNN(color)). 
    }
    \label{fig:result_loss_reduction_L3D32}
\end{figure}

Fig.~\ref{fig:result_loss_reduction_L3D32} is the scatter plot of the $(\bar r_{train},  \bar r_{loss})$, $L=3$ and $D=32$.  
In this case, two HIV dataset plus one case for QM9 reported the increase of the training loss.
For other 13 pairs, the GWM successfully reduce the training losses. 
It is remarkable that all 16 plots have positive test loss reduction rates: namely, the generalization performances are improved by the GWM in this choice of the hyperparameters. 

\begin{figure}[h]
\centering
    \includegraphics[width=73mm]{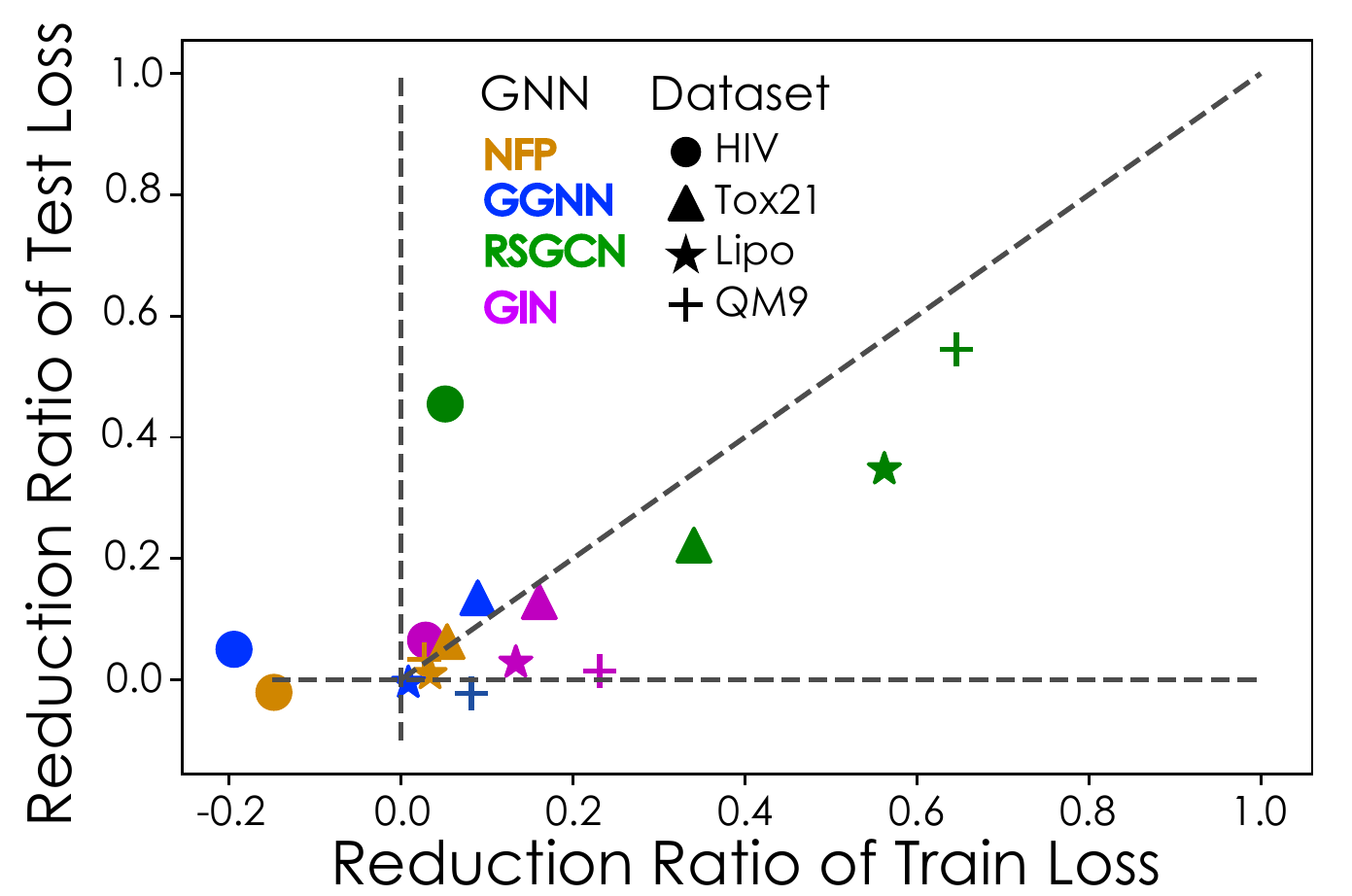}
    \caption{Train (horizontal) and test (vertical) loss reduction ratios on various pairs of GNN models and datasets. $L=4$, $D=100$. 
    Each plot presents the rational train/test loss reductions induced by the GWM attachment for a specific pair of (dataset(symbol), GNN(color)). 
    }
    \label{fig:result_loss_reduction_L4D100}
\end{figure}

Fig.~\ref{fig:result_loss_reduction_L4D100} is the scatter plot of the $(\bar r_{train},  \bar r_{loss})$, $L=4$ and $D=100$.    
In this case, only two cases of the HIV datasets record the negative $\bar r_{train}$ values. 
For other 14 pairs, the GWM successfully reduce the training losses and 13 out of these 14 pairs have positive $\bar r_{test}$s. 

All models were trained for 30 epochs. 

\subsection{Section 4.4: the full comparison}
For the experiments in the Section 4.3 (Table 2 and 3 in the main manuscript), 
we employed the Bayesian optimization to tune $L$ and $D$ for each combinations of a dataset and a GNN model. 
The Bayesian optimization (BO) trials were conducted by the Optuna library, with 200 sampling (searches) for each combination. 
Ranges of the BO search is: $2 \leq L \leq 8$, $4 \leq D \leq 512$. 

Chosen $L$s and $D$s are presented in the Table~\ref{tab:app_hyprms}. 

For the QM9 dataset, we trained the models for 50 epochs. 
For the HIV and the Tox21 dataset, we trained the models for 100 epochs. 
For the LIPO dataset, we trained the models for 200 epochs. 

Tables~\ref{tab:app_result_lipo_qm9_BO} and~\ref{tab:app_result_hiv_tox21_BO} are the full lists of the main comparison experiments in the main manuscript, with the standard deviation values in parentheses. 

\begin{table}[t]
    \centering
    \small
    \begin{tabular}{c|c||c|c|c|c|c|c}
         Dataset & GNN model & NFP & WeaveNet & RGAT & GGNN & RSGCN & GIN \\ \hline 
         \multirow{2}{*}{LIPO} & vanilla GNN & (4, 232) & (3, 50) & (4, 9) & (4, 32) & (5, 19) & (6, 19)  \\
                               & $+$Simple Supernode & (2, 71) & (3, 9) & (3, 8) & (4, 18) & (2, 27) & (3, 14) \\
                               & $+$NoGate GWM & (2, 65) & (2, 14) & (3, 12) & (5, 30) & (2, 15) & (6, 9) \\
                               & $+$Proposed GWM & (5, 231) & (4, 15) & (5, 19) & (6, 127) & (4, 22) & (2, 26) \\
                               \hline
         \multirow{2}{*}{QM9} & vanilla GNN & (5, 86) & (3, 22) & (4, 40) & (5, 71) & (4, 100) & (4, 250) \\
                                & $+$Simple Supernode & (5, 44) & (3, 111) & (6, 34) & (5, 60) & (5, 69) & (3, 22) \\
                               & $+$NoGate GWM & (4, 48) & (5, 250) & (4, 31) & (6, 50) & (4, 48) & (4, 34) \\
                              & $+$Proposed GWM & (5, 72) & (3, 104) & (4, 156) & (8, 50) & (4, 36) & (4, 23) \\ \hline
         \multirow{2}{*}{HIV} & vanilla GNN & (6, 213) & (2, 65) & (3, 28) & (6, 29) & (4, 57) & (2, 72) \\
                                & $+$Simple supernode & (3, 30) & (3, 39) & (2, 9) & (4, 54) & (4, 255) & (5, 93) \\
                               & $+$NoGate GWM & (4, 46) & (3, 20) & (2, 12)  & (2, 51) & (3, 27) & (3, 60) \\
                               & $+$Proposed GWM & (3, 200) & (3, 92) & (3, 23) & (8, 135) & (3, 106) & (8, 38) \\ \hline
         \multirow{2}{*}{Tox21} & vanilla GNN & (3, 204) & (5, 90) & (3, 19) & (6, 36) & (5, 70) & (5, 103) \\
                               & $+$Simple supernode & (5, 129) & (6, 31) & (2, 36) & (6, 136) & (5, 119) & (5, 157)\\
                        & $+$NoGate GWM & (2, 123) & (2, 157) & (3, 43) & (5, 79) & (4, 31) & (6, 117)\\
                    & $+$Proposed GWM & (3, 106) & (4, 19) & (3, 37) & (7, 48) & (8, 32) & (6, 102)
    \end{tabular}
    \caption{Hyperparameters $L$ and $D$ for the experiment in Section 4.4. The format of the table cells is: $(L, D)$.  }
    \label{tab:app_hyprms}
\end{table}

\newpage
\begin{landscape}
\begin{table}[t]
    \centering
    \small
    \begin{tabular}{c|c||c|c|c|c|c|c}
         Dataset & GNN model & NFP & WeaveNet & RGAT & GGNN & RSGCN & GIN\\ \hline 
         \multirow{2}{*}{LIPO} & vanilla GNN & .677 (.040) & 1.19 (.327) & .753 (.045) & .582 (.022) & .801 (.014) & .844 (.026)\\
                               & $+$Simple Supernode & 
                               .693 (.027) & 
                               \textbf{1.01} (.208) & 
                               \textbf{.740} (.032) & 
                               .604 (.027) & 
                               \textbf{.775} (.011) & 
                               .819 (.023) \\
                               & $+$NoGate GWM & 
                               \textbf{.675} (.017) & 
                               \textbf{.721} (.225) & 
                               \textbf{.688} (.013) & 
                               \textbf{.576} (.022) & 
                               \textbf{.787} (.048) & 
                               .847 (.048) \\
                               & $+$Proposed GWM & 
                               \textbf{.672*} (.040) & \textbf{.688*} (.105) & \textbf{.659*} (.016) & \textbf{.569*} (.022) & \textbf{.752*} (.014) & \textbf{.784*} (.012)\\
                               \hline
         \multirow{2}{*}{QM9} & vanilla GNN & 6.16 (.231) & 6.38 (.289) & 8.96 (.192) & 4.92 (.145) & 15.2 (1.05) & 14.0 (2.47) \\
                                & $+$Simple Supernode & 
                                7.68 (.316) & 
                                \textbf{5.51} (.248) & 
                                9.00 (.399) & 
                                5.41 (.110) & 
                                14.6 (1.71) & 
                                \textbf{11.5*} (.755) \\
                               & $+$NoGate GWM & 
                               6.84 (.348) & 
                               \textbf{5.40*} (.238) & 
                               9.21 (.368) & 
                               5.52 (.344) & 
                               \textbf{12.5} (.564) & 
                                \textbf{12.9} (1.32) \\
                              & $+$Proposed GWM & 
                              6.64* (.360) & 
                              \textbf{5.90} (.248) & 
                              \textbf{8.39*} (.219) & 
                              \textbf{4.88*} (.231) & \textbf{11.9*} (1.48) & \textbf{11.8} (1.21) 
    \end{tabular}
    \caption{MAEs on the LIPO dataset and QM9 dataset. The number of layers and the dimension of the feature vectors are defined via Bayesian Optimization for each method and each dataset. Averages (standard deviations) over 10 random runs. Smaller values are better. }
    \label{tab:app_result_lipo_qm9_BO}
\end{table}
\begin{table}[t]
    \centering
    \small
    \begin{tabular}{c|c||c|c|c|c|c|c}
         Dataset & GNN model & NFP & WeaveNet & RGAT & GGNN & RSGCN & GIN\\ \hline 
         \multirow{2}{*}{HIV} & vanilla GNN & .724 (.017) & .670 (.020) & .707(.039) & .746 (.018) & .746 (.011) & .729 (.020)\\
                                & $+$Simple supernode & 
                                .707 (.023) & 
                                \textbf{.676} (.031) & 
                                .704 (.019) & 
                                \textbf{.764*} (.014) & 
                                .728 (.009) & 
                                .729 (.025) \\
                               & $+$NoGate GWM & 
                               .714 (.018) & 
                               \textbf{.680} (.043) & 
                               .726 (.024) & 
                               .744 (.006) & 
                               .742 (.019) & 
                               \textbf{.739} (.012) \\
                               & $+$Proposed GWM & 
                               \textbf{.731*} (.020) & 
                               \textbf{.681*} (.010) & \textbf{.748*} (.019) & \textbf{.762} (.016) & \textbf{.758*} (.022) & 
                               \textbf{.755*} (.009) \\ \hline
         \multirow{2}{*}{Tox21} & vanilla GNN & .763 (.004) & .710 (.029) & .744 (.008) & .764 (.009) & .760 (.005) & .740 (.007)\\
                        & $+$Simple supernode & \textbf{.770} (.008) & \textbf{.750} (.020) & \textbf{.787*} (.005) & \textbf{.790} (.006) & \textbf{.770*} (.006) & \textbf{.763} (.006) \\
                        & $+$NoGate GWM & 
                        \textbf{.775*} (.007) & 
                        \textbf{.764} (.011) & 
                        \textbf{.786} (.009) & 
                        \textbf{.792*} (.008) & 
                        .759 (.007) & 
                        \textbf{.766} (.008) \\
                        & $+$Proposed GWM & 
                        \textbf{.769} (.007) & 
                        \textbf{.767*} (.021) & 
                        \textbf{.787*} (.009) & 
                        \textbf{.785} (.005) & 
                        \textbf{.769} (.006) & 
                         \textbf{.768*} (.007) 
    \end{tabular}
    \caption{ROC-AUCs on the HIV dataset and Tox21 dataset. The number of layers and the dimension of feature vectors are defined via Bayesian Optimization for each method and each dataset. Averages (standard deviations) over 10 random runs. Larger values are better. }
    \label{tab:app_result_hiv_tox21_BO}
\end{table}
\end{landscape}

\end{document}